\documentclass{article}

\usepackage[preprint]{jmlr2e}

\usepackage[utf8]{inputenc} % allow utf-8 input
\usepackage[T1]{fontenc}    % use 8-bit T1 fonts
\usepackage{hyperref}       % hyperlinks
\usepackage{url}            % simple URL typesetting
\usepackage{amsfonts}       % blackboard math symbols
\usepackage{nicefrac}       % compact symbols for 1/2, etc.
\usepackage{microtype}      % microtypography
\usepackage{xcolor}         % colors
\usepackage{graphicx}
\usepackage{subcaption}
\usepackage{stfloats}

\newcommand{\methodname}{TExplain}

\title{TExplain: Explaining Learned Visual Features via Pre-trained (Frozen) Language Models}

\author{\\ \name Saeid A. Taghanaki* \email saeid.asgari.taghanaki@autodesk.com\\
       \addr Autodesk Research\\
       \AND
       \name Aliasghar Khani* \email aliasghar\_khani@sfu.ca\\
       \addr Autodesk Research\\
       \AND
       \name Ali Saheb Pasand \email asahebpa@uwaterloo.ca\\
       \addr Autodesk Research\\
       \AND
       \name Amir Khasahmadi \email amir.khasahmadi@autodesk.com \\ 
       \addr Autodesk Research\\
       \AND
       \name Aditya Sanghi \email aditya.sanghi@autodesk.com\\ 
       \addr Autodesk Research\\
       \AND
       \name Karl D.D. Willis \email karl.willis@autodesk.com\\ 
       \addr Autodesk Research\\
       \AND
       \name Ali Mahdavi-Amiri \email ali\_mahdavi-amiri@sfu.ca\\
       \addr Simon Fraser University}

\begin{document}

\maketitle
\def\thefootnote{*}\footnotetext{These authors contributed equally to this work.}

\begin{abstract}
Interpreting the learned features of vision models has posed a longstanding challenge in the field of machine learning. To address this issue, we propose a novel method that leverages the capabilities of language models to interpret the \textit{learned features} of pre-trained image classifiers.
Our method, called \methodname{}, tackles this task by training a neural network to establish a connection between the feature space of image classifiers and language models. Then, during inference, our approach generates a vast number of sentences to explain the features learned by the classifier for a given image. These sentences are then used to extract the most \textit{frequent words}, providing a comprehensive understanding of the learned features and patterns within the classifier.
Our method, for the first time, utilizes these frequent words corresponding to a visual representation to provide insights into the decision-making process of the independently trained classifier, enabling the detection of spurious correlations, biases, and a deeper comprehension of its behavior. To validate the effectiveness of our approach, we conduct experiments on diverse datasets, including ImageNet-9L and Waterbirds. The results demonstrate the potential of our method to enhance the interpretability and robustness of image classifiers.
\end{abstract}

\section{Introduction}

%% opening 

Discriminative visual models have achieved impressive results across a broad range of tasks. However, their decision-making process remains challenging to interpret~\citep{adebayo2018sanity}. This lack of transparency hinders their use in real-world scenarios where interpretability is crucial. Providing explanations for models enables practitioners to comprehend how they operate, detect and rectify errors, and influence their decisions. However, most existing interpretability tools have been criticized for generating explanations that may contain considerable errors, based on computational or qualitative user-study evidence, and should be used with caution~\citep{adebayo2018sanity,chu2020visual,poursabzi2021manipulating,kindermans2019reliability,srinivas2020rethinking,alqaraawi2020evaluating}. Regardless of providing error-prone explanations, such explanation-based approaches are only able to indicate the most crucial input variables for a specific prediction, while being unable to identify the predominate features in a visual representation vector. In this work, we are interested in deciphering the nature of these features and explore the extent to which they are embedded within a visual representation vector which a classifier uses to make a prediction. To achieve this, we leverage the capabilities of language models to bridge the gap between the latent semantics encoded within visual representation vectors and their interpretation in a human-understandable form.

Over the past few years, there has been remarkable progress on langugage models, which have demonstrated extraordinary capabilities in various natural language processing tasks. The introduction of models like BERT~\citep{devlin2018bert} and GPT variants~\citep{brown2020language} and subsequent advancements have showcased the immense potential of these models in generating coherent and contextually relevant text. Alongside language processing, these models have also found applications in diverse domains such as image classification when they are coupled with vision encoders~\citep{radford2021learning}, machine translation~\citep{vaswani2017attention}, and question-answering~\citep{brown2020language}. Despite the significant advancements made in language modeling, there remains a need to explore the potential of these models in interpreting and explaining complex systems, particularly in the context of independently trained image classifiers. This paper aims to address this gap and investigate the potential of leveraging language models for interpreting independently trained image classifiers, shedding light on their interpretability and providing valuable insights into their decision-making processes.

In this paper, our goal is to address the question of \textit{how to leverage a trained (frozen) language model to translate the learned visual features of an independently trained and frozen classifier into textual explanations}. We aim to decipher the incomprehensible feature vectors into easily understandable textual explanations, enabling us to assess if the learned feature vectors capture meaningful information.

We present a novel method called \methodname{} that utilizes a pre-trained language model to analyze the learned representations of image classifiers. \methodname{} generates textual explanations comprising prominent descriptive terms that correspond with the visual features of the input. This is achieved by generating a vast set of highly probable sentences for each visual feature vector and then identifying the most frequently occurring descriptive words for each visual representation using \methodname{}. Our method offers a new perspective for interpreting and understanding the representations learned by vision models.

\begin{figure}[t]
    \centering
     \includegraphics[width=0.7\textwidth]{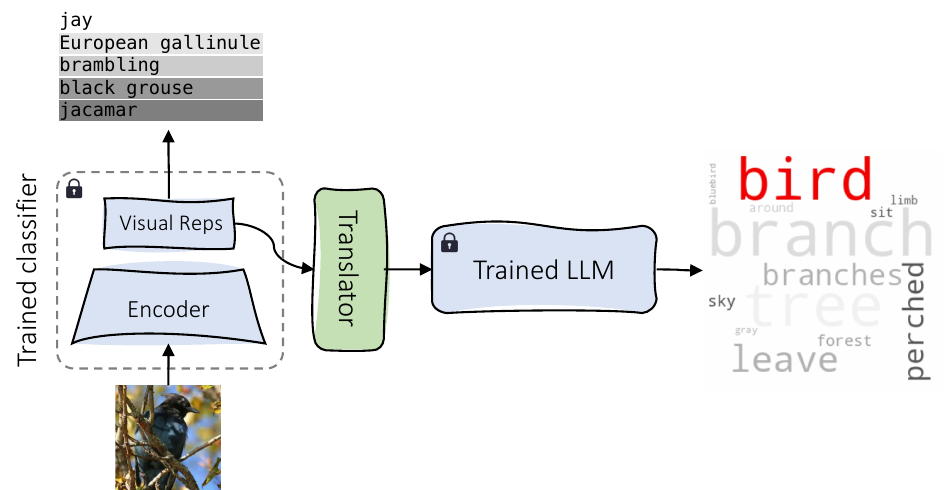}
     \caption{\methodname{} projects learned visual representations of a frozen image classifier onto a space that an independently trained language model can interpret. Using a large number of generated sentence samples along with the visual representation, \methodname{} produces a word cloud for each visual representation. Blue and green refers to frozen and trainable parameters, respectively. The category of the feature representation is highlighted in red, while other captured features are shown in gray. The font size of each word indicates the strength of its corresponding feature.}
     \label{fig:method} 
 \end{figure}

Figure~\ref{fig:method} visualizes how \methodname{} can be used to discover the most frequent words from representations learned by an image classifier. The challenge here is to convert visual representations into language model-processable inputs. Therefore, \methodname{} trains a small multilayered perceptron to map visual features to the space of language models.

%% contributions
To the best of our knowledge, this is the first work to present a technique to encode learned visual features to textual explanations. Our contributions are summarized as follows:

\begin{enumerate}
    \item We introduce \methodname{}, a novel approach that utilizes language models to explain the \textit{learned features} of independently trained and frozen image classifiers.
    \item We demonstrate that by performing minor feature translation, it is possible to generate explanations for frozen image classifiers using pre-trained language models.
    \item Through empirical analysis, we validate the effectiveness of \methodname{} in identifying spurious features within a specific class.
    \item We illustrate the practical application of \methodname{} by showcasing how it can be leveraged to mitigate spurious correlations within a dataset.
\end{enumerate}

\section{Method}

Our method aims to elucidate the characteristics of image classifiers by leveraging pre-trained language models. To accomplish this, our architecture comprises three key components: a pre-trained frozen image classifier, a trainable translator network, and a pre-trained language model. An overview of the pipeline is depicted in Figure \ref{fig:method}.
During the training phase, our approach involves training a translator network to establish a connection between the features of the frozen image classifier and the pre-trained frozen language model, utilizing pairs of (image, caption).
This enables us, during the \textit{inference} stage, to provide explanations regarding the learning process of the image classifier for a given image by extracting the most frequent words among all its corresponding explanatory sentences.

\subsection{Pre-trained Image Classifier}
The primary component of our approach is the image classifier whose features we aim to interpret.

The input to the classifier is image $I \in \mathbb{R}^{H \times W \times C}$ where $H$, $W$, and $C$ are height, width, and the number of channels of the input image, respectively.
To obtain the feature $Z$ that we aim to interpret, we pass the image $I$ through the pretrained (frozen) image classifier $enc(.)$. The resulting embedding $Z$ is obtained as $Z = enc(I)$ which in our case is the penultimate layer (before the classification layer).
 
While the choice of image classifier can vary, we specifically consider ViT \citep{dosovitskiy2020image} due to its widespread usage. ViT's architecture allows for efficient processing of large-scale image datasets and robust feature extraction. It splits the image into $P$ patches, adds a learnable token to the patch or token embeddings, and produces a $(P+1) \times D$ matrix, where $D$ represents the embedding dimension of each token. Hence, in the case of ViT, the feature $Z$ can be expressed as $Z \in \mathbb{R}^{(P+1) \times D}$.

\subsection{Training the Translator Network}
The embedding \(Z\) generated by the image encoder represents the key characteristics captured by the classifier from the input image. During inference, our goal is to interpret this feature vector. To achieve this, we aim to transform the representation into a human-understandable description using natural language. This involves mapping the embeddings generated by the classifier's encoder to the embedding space of a language model.

Concretely, \(Z\) is flattened into a 1-dimensional vector \(Z_{in} \in \mathbb{R}^{1 \times ((P+1)*D)}\). This vector is then passed through a translator network, denoted as \(t(\cdot)\), to obtain \(Z_{mapped} = t(Z_{in})\). Here, \(Z_{mapped}\) shares the same size as the input of the text decoder of the language model. The translator is the only component in our framework that requires training and has a simple linear multi-layer perceptron (MLP) architecture with batch normalization.

To train \(t(\cdot)\), we utilize image-sentence pairs (\(I, S\)). \(Z_{in}\) is calculated given \(I\) as the input to the classifier, and \(Z_{mapped}\) is learned by minimizing the language model loss. The language model loss is defined as the cross-entropy loss between \(S_{gen}\) and the ground truth sentence \(S\). To generate \(S_{gen}\), \(Z_{mapped}\) is passed to the decoder (\(dec(\cdot)\)) of a pre-trained frozen language model. \(S_{gen} = dec(Z_{mapped})\).

Once the translator network is trained, during inference, \(S_{gen}\) serves as an explanation of the visual embedding captured by the frozen image classifier. This sheds light on its underlying features and patterns.

% The embedding generated by the image encoder, $Z$, represents the key characteristics captured by the classifier from  the input image. During inference, we want to interpret this feature vector.  Our objective is then to transform this representation into a human-understandable description using natural language. To accomplish this, we need to map the embeddings generated by the encoder of the classifier to the embedding space of a language model. 
% Concretely, $Z$ is flattened into a 1-dimensional vector $Z_{in} \in \mathbb{R}^{1 \times ((P+1)*D)}$, then passed through a translator network, $t(.)$,  to obtain $Z_{mapped}=t(Z_{in})$, where $Z_{mapped}$ is the same size as the input of the text decoder of the language model. The translator is the only part in our framework that needs to be trained and has a simple multi-layer perceptron (MLP) architecture.

% In order to train $t(.)$, we use image-sentence pairs ($I, S$), such that $Z_{in}$ is calculated given $I$ as input to the classifier, and $Z_mapped$ is learned by minimizing the language model loss, which is the cross entropy loss between $S_{gen}$ and the ground truth sentence $S$. To generate $S_{gen}$, $Z_mapped$ is passed to the decoder, $dec(.)$, of a pre-trained frozen language model, which can be of any choice, similar to the image classifier. sentence $S_{gen}=dec(Z_{mapped})$.

% After training the translator network, during inference, $S_{gen}$, serves as an explanation of the visual embedding captured by the frozen image classifier, shedding light on its underlying features and patterns.

\subsection{Identifying Dominant Words by Sampling}

To minimize potential noise and enhance the reliability of the generated sentences from the language model, we employ Nucleus Sampling \citep{holtzman2019curious}. This technique allows us to sample a set of $N$ sentences, denoted as $\{S_{gen}^{i}\}_{i=1}^N$. By removing the less frequently occurring words, we construct a word cloud based on the dominant words extracted from the set of sentences $\{S_{gen}^{i}\}_{i=1}^N$. This word cloud visually represents the prominent features within the visual embedding of the frozen classifier. By focusing on these dominant words, we gain insights into the key characteristics and attributes captured by the classifier's visual representation. The word cloud serves as a concise and informative summary of the significant features present in the embedding space of the image encoder. It worth noting that focusing on frequent dominant words reduces the effect of \textit{hallucinations}~\citep{maynez2020faithfulness} imposed by language models, as those words appear in majority of the generated sentences given a feature vector.

\section{Implementation details}

\textbf{Models.} While we acknowledge that alternative variants of the main models can be substituted, we have chosen to employ widely recognized and popular models for the sake of simplicity. Specifically, we utilized the pretrained ViT-base model~\citep{wu2020visual} as our image classifier. This model incorporated 577 tokens and processed input images at a resolution of $384 \times 384$ pixels. For the language model, we utilized the pre-trained BERT-base model~\citep{devlin2018bert}, featuring 12 layers and 12 attention heads. Regarding the translator component, we utilize a straightforward architecture consisting of a three-layered linear MLP with batch normalization. In the case of ViT, its input size is \(577 \times 768\) and outputs the same dimension, which is then reshaped to match the input of \(dec(\cdot)\).

\textbf{Sampling Explanations.} Using nucleus sampling, we sample 1000 sentences from each visual representation. Hence, when generating class-level explanations, we will have $N \times 1000$, where $N$ represents the number of samples in that class. To maintain coherence, we set the cumulative probability threshold to 0.95. Additionally, we define the minimum and maximum length of the generated sentences to be 20 and 30 words, respectively. This sampling strategy allows us to capture a range of explanations that effectively convey the salient features present in the visual representations.

\textbf{Data.} We used a comprehensive dataset comprising a total of 14 million data points. This dataset encompassed COCO~\citep{lin2014microsoft} and Visual Genome~\citep{krishna2017visual} which come with human annotations, as well as three web datasets, including Conceptual Captions and Conceptual 12M~\citep{changpinyo2021conceptual}, and SBU captions~\citep{ordonez2011im2text}. We trained the translator using all the data, excluding the COCO dataset, for 20 epochs. Subsequently, we fine-tuned the translator using the COCO dataset for an additional 5 epochs. Throughout the training process, a batch size of 512 was employed.

\section{Experiments}

In this section, we present the experiments conducted to investigate the capabilities of \methodname{}, our proposed language model-based technique. We commence with a straightforward and intuitive experiment on an altered version of the Cats vs Dogs dataset~\citep{elson2007asirra}, where our goal is to train an image encoder to learn co-occurring features within images. Through this experiment, we aim to evaluate the effectiveness of our \methodname{} in capturing relevant and meaningful information from the trained model. Subsequently, we delve into an analysis of shortcuts on the Background Challenge dataset~\citep{xiao2020noise} and spurious correlations on the Waterbirds dataset~\citep{sagawa2019distributionally} using \methodname{} as a means to mitigate these correlations.

\subsection{Faithfulness Test 1: Verifying that \methodname{} Picks up Relevant Features}

\begin{figure}[h]
     \centering
     \includegraphics[width=\textwidth]{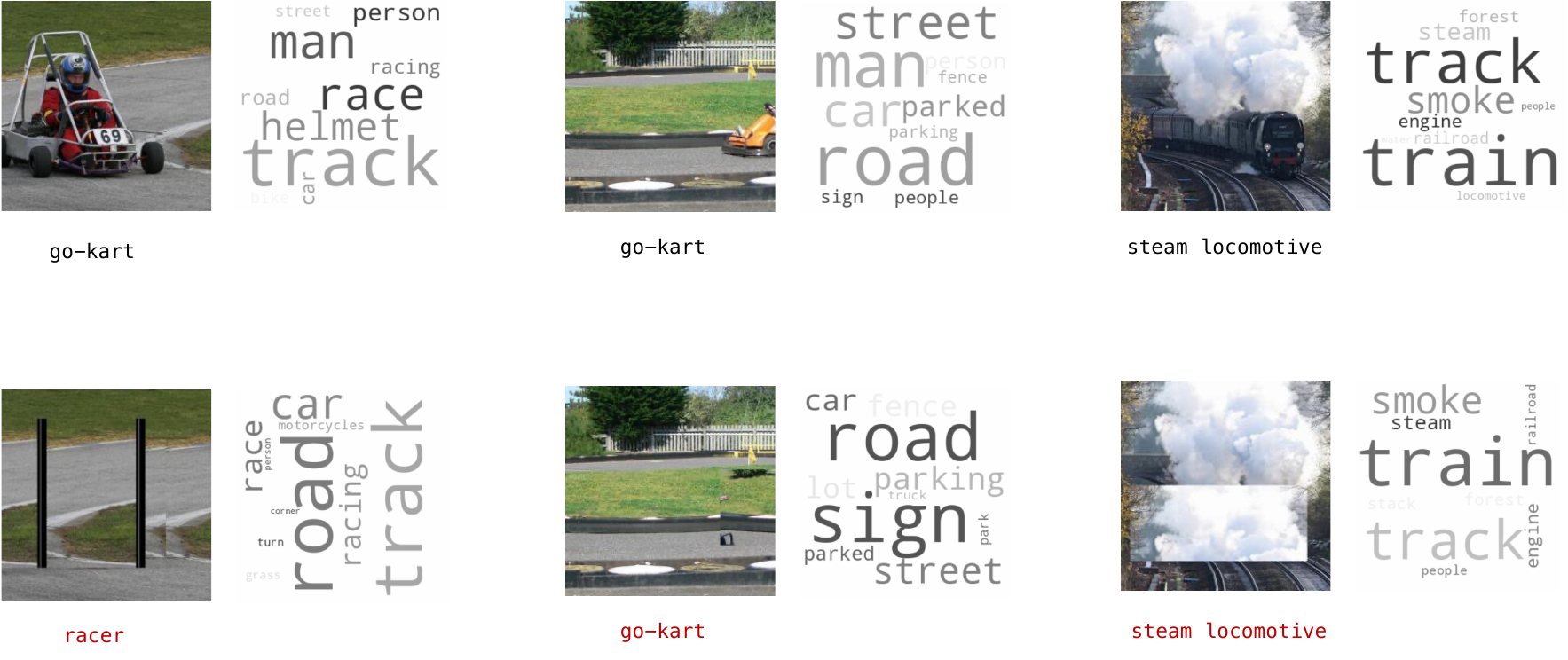}
     \caption{Class predictions and corresponding word clouds generated by \methodname{} for both the original (top) and Only-BG-T (bottom) samples extracted from the Background Challenge dataset. The class predictions are displayed below each image.}
     \label{fig:bg_exp2}
 \end{figure}

In this experiment, we assess an Imagenet classifier's performance using images in which the foreground is hidden (Only-BG-T from the Background Challenge datastet). Our hypothesis is that if the classifier consistently assigns the same label to an image, regardless of whether the foreground is visible or concealed, it indicates the presence of potentially misleading correlations between different regions of the image and the classifier's predictions. Consequently, the \methodname{} should effectively bring attention to these correlations. In Figure~\ref{fig:bg_exp2}, we present samples from the Background challenge dataset, illustrating instances where the classifier consistently assigns the same label to the images, even when the foreground is concealed. To shed light on the underlying associations, we utilize \methodname{} to generate word clouds from frequent words for each sample. These word clouds effectively highlight the correlated shortcuts present in each image. For instance, we observe a notable co-occurrence of "smoke," "train," and "track," which the classifier relies on as shortcuts for the \texttt{steam locomotive} category. This visualization further emphasizes the classifier's dependence on these spurious correlations that \methodname{} identifies. 

\subsection{Faithfulness Test 2: Human Analysis}
\begin{figure}[h]
     \centering
     \includegraphics[width=0.6\textwidth]{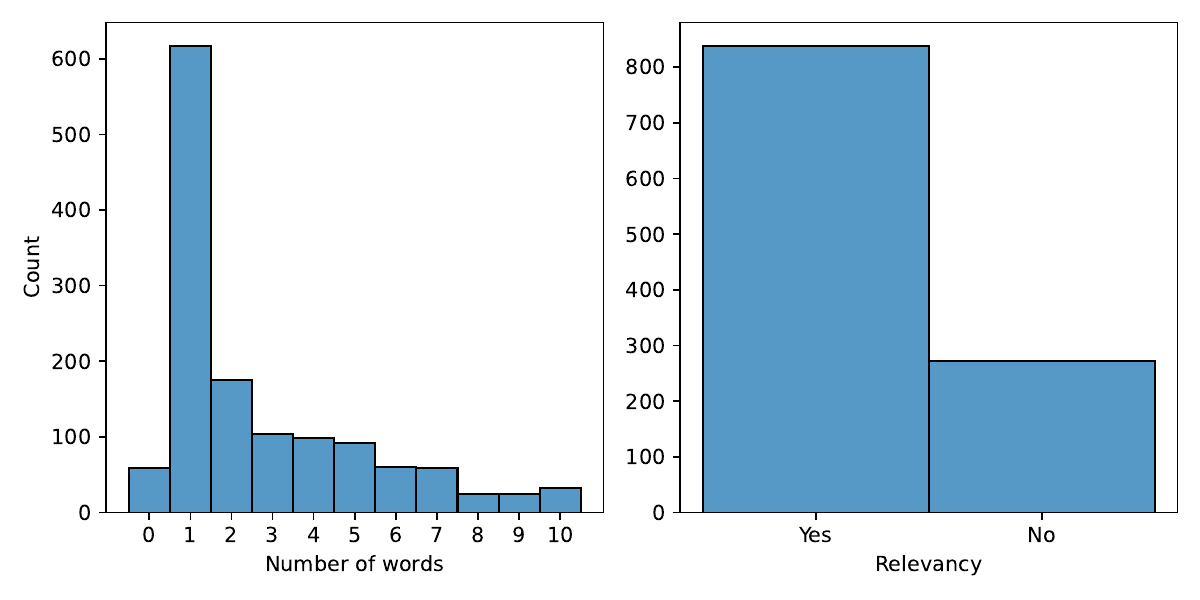}
     \caption{Human analysis experiments were conducted to validate that \methodname{} generates meaningful explanations. The left side shows the count of relevant words, while the right side illustrates whether the most frequent word produced by \methodname{} is relevant to the classifier's top prediction.}
     \label{fig:human_analysis}
 \end{figure}

We conducted two separate human analysis experiments on three different from the background challenge (ImageNet-9) dataset: 
a) Raters were presented with a set of the 10 most frequent words produced by \methodname{} for each feature vector alongside the corresponding image. They were then asked to indicate how many words accurately matched the content of the image. b) In another experiment, raters were provided with the classifier's top prediction for an image and the top word in the most frequent list generated by \methodname{}. Raters were instructed to determine the relevance of these two words, providing 'yes' or 'no' responses. The results of these experiments are illustrated in Figure~\ref{fig:human_analysis}. As depicted in Figure~\ref{fig:human_analysis} (left), the raters consistently identified matching words with the image among the ten most frequent words. Similarly, Figure~\ref{fig:human_analysis} (right) demonstrates that the classifier's top prediction frequently aligns with the most frequent word generated by \methodname{}.

\subsection{Faithfulness Test 3: Verifying using Generative Model's Latent Space}

In this experiment, our goal is to assess the ability of the \methodname{} technique to emphasize prominent features within the latent space of Stable Diffusion (SD)~\citep{rombach2022high}. Our rationale is based on the fact that SD generates an image from a latent vector, meaning this vector should encapsulate sufficient information to create a corresponding image. Our investigation aims to confirm the correlation between the textual features we extract from SD's latent space (using \methodname{}) and the features that can be derived from the image using standard multi-modal models. To pinpoint the primary objects or features within the generated image, we utilize the BLIP method \citep{li2022blip} to generate descriptive captions for the output image. We expect  our \methodname{}'s explanations within the latent space to align with BLIP-generated captions in the output space. To asses this we start by selecting a specific category, for example, "kitchen." Using corresponding category captions from the COCO dataset as prompts, we generate 100 images for each prompt using the SD model. At the same time, we employ BLIP to generate captions for these newly created images. During this process, we also extract latent features before generating each image. These latent features originate from the final step of the SD model, just before they are passed through the decoder of the variational auto-encoder to create an image. Subsequently, we take these latent features and process them through our \methodname{} model, resulting in explanations situated within the latent space.

\begin{figure}
    \centering
    
    \begin{subfigure}{0.25\textwidth}
        \includegraphics[width=\linewidth]{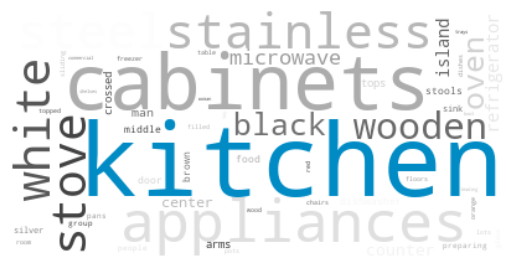}
        % \caption{bus}
    \end{subfigure}
    \begin{subfigure}{0.25\textwidth}
        \includegraphics[width=\linewidth]{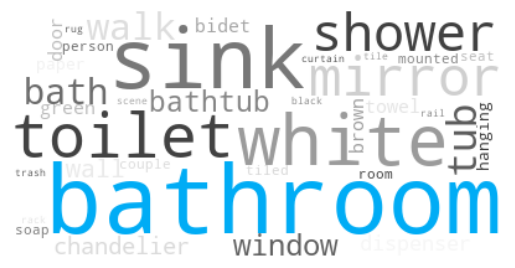}
        % \caption{Subfigure 2}
    \end{subfigure}
    \begin{subfigure}{0.25\textwidth}
        \includegraphics[width=\linewidth]{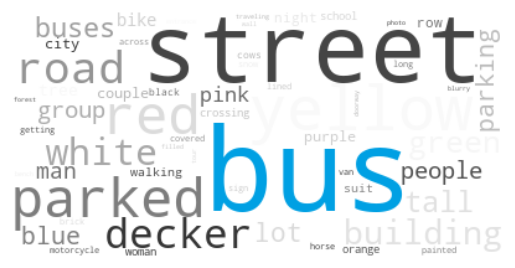}
        % \caption{Subfigure 3}
    \end{subfigure}

    \vspace{0.5cm} 

    \begin{subfigure}{0.25\textwidth}
        \includegraphics[width=\linewidth]{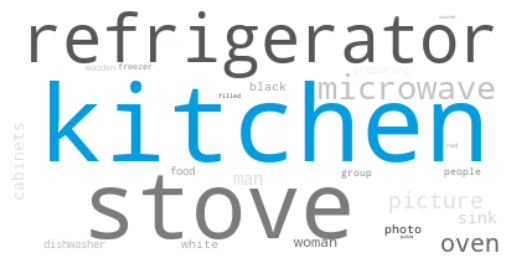}
        % \caption{Subfigure 4}
    \end{subfigure}
    \begin{subfigure}{0.25\textwidth}
        \includegraphics[width=\linewidth]{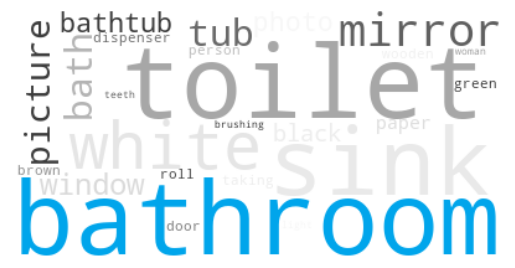}
        % \caption{Subfigure 5}
    \end{subfigure}
    \begin{subfigure}{0.25\textwidth}
        \includegraphics[width=\linewidth]{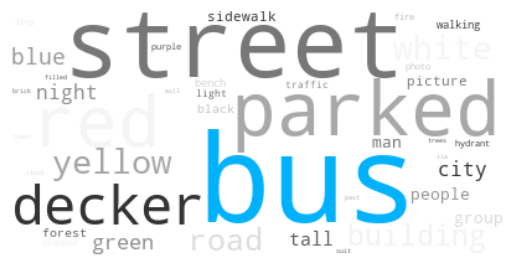}
        % \caption{Subfigure 6}
    \end{subfigure}
    
    \caption{Wordclouds based on image captioning (top) with BLIP and feature captioning (bottom) using \methodname{} for categories \texttt{kitchen, bathroom,} and \texttt{bus}.}
    \label{fig:stable_diff}
\end{figure}

We then proceed to create two word clouds for each category: one based on the image captions generated by BLIP and another derived from the latent space explanations produced by \methodname{}. In Figure~\ref{fig:stable_diff}, we present a visual comparison between the word clouds generated by BLIP (at the top) and those generated by \methodname{} (at the bottom) for three distinct categories, namely "kitchen," "bathroom," and "bus." As shown in the figure, the explanations provided by \methodname{} contains a similar distribution of objects and categories when compared to BLIP's captions. This observation underscores the ability of \methodname{} to generate faithful explanations that align with the features present in the output space.'

We then extend this to 26 object categories and compute image captioning metrics such as ROUGE~\citep{lin2004rouge} and METEOR~\citep{banerjee2005meteor}, as well as cosine similarity between the sentence embeddings obtained using BERT for both \methodname{} and BLIP. We report these results in Table~\ref{tab:texplain_blip}. Notably, the average cosine distance for \methodname{} and BLIP across all the categories is approximately 0.85, indicating that \methodname{} identifies learned features.

\begin{table}[h!]
\setlength\tabcolsep{4pt}
\centering
\caption{Quantifying the relevancy of textual feature-based explanations by \methodname{} and image-based captions using BLIP.}
\begin{tabular}{lccc}
\cline{2-4}
                               & Cosine similarity & ROUGE & METEOR \\ \hline
Scores & 0.845             & 40.39 & 38.95  \\ \hline
\end{tabular}
\label{tab:texplain_blip}
\end{table}

\subsection{Detecting Potential Shortcuts/Spurious Correlations}

\textbf{ImageNet-9L.} In an effort to extend our evaluation, we conducted a comprehensive analysis of \methodname{} using the Background Challenge dataset~\citep{xiao2020noise}. The Background Challenge dataset is publicly accessible and comprises test sets derived from ImageNet-9~\citep{deng2009imagenet}, containing diverse foreground and background signals. Its primary objective is to assess the extent to which deep classifiers depend on irrelevant features for image classification.

\begin{figure}[h!]
     \centering
     \includegraphics[width=0.8\textwidth]{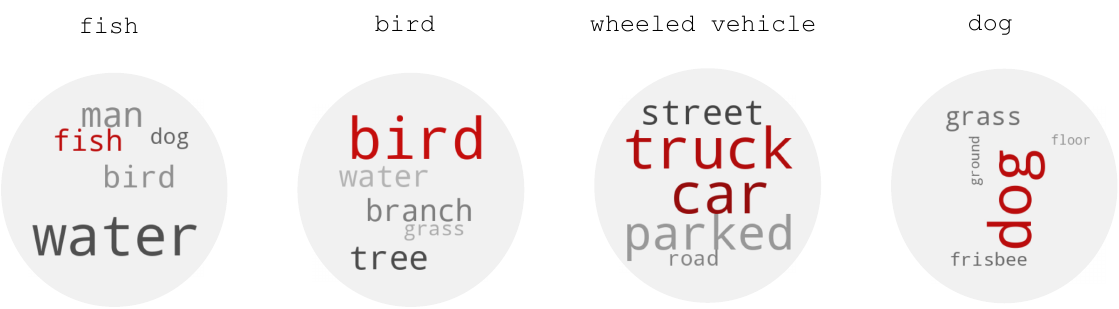}
     \caption{Word clouds generated from \methodname{} explanations for the Background Challenge Dataset categories. Red represents the detected features related to the main category.}
     \label{fig:bg_ch_wc}
 \end{figure}

To analyze the prominent features within each category of this dataset, we employed \methodname{} to generate word clouds for all categories. Figure~\ref{fig:bg_ch_wc} showcases the outcomes of this process. In the \texttt{wheeled vehicle} category, dominant features such as "street", "truck", and "car" emerged prominently. Conversely, in the \texttt{fish} category, the primary feature observed was "water", which exhibited an even stronger influence than the \texttt{fish} itself. These findings strongly indicate that the classifier is more likely to rely on shortcut features rather than the genuine object features.

\begin{figure}[ht]
     \centering
     \includegraphics[width=0.7\textwidth]{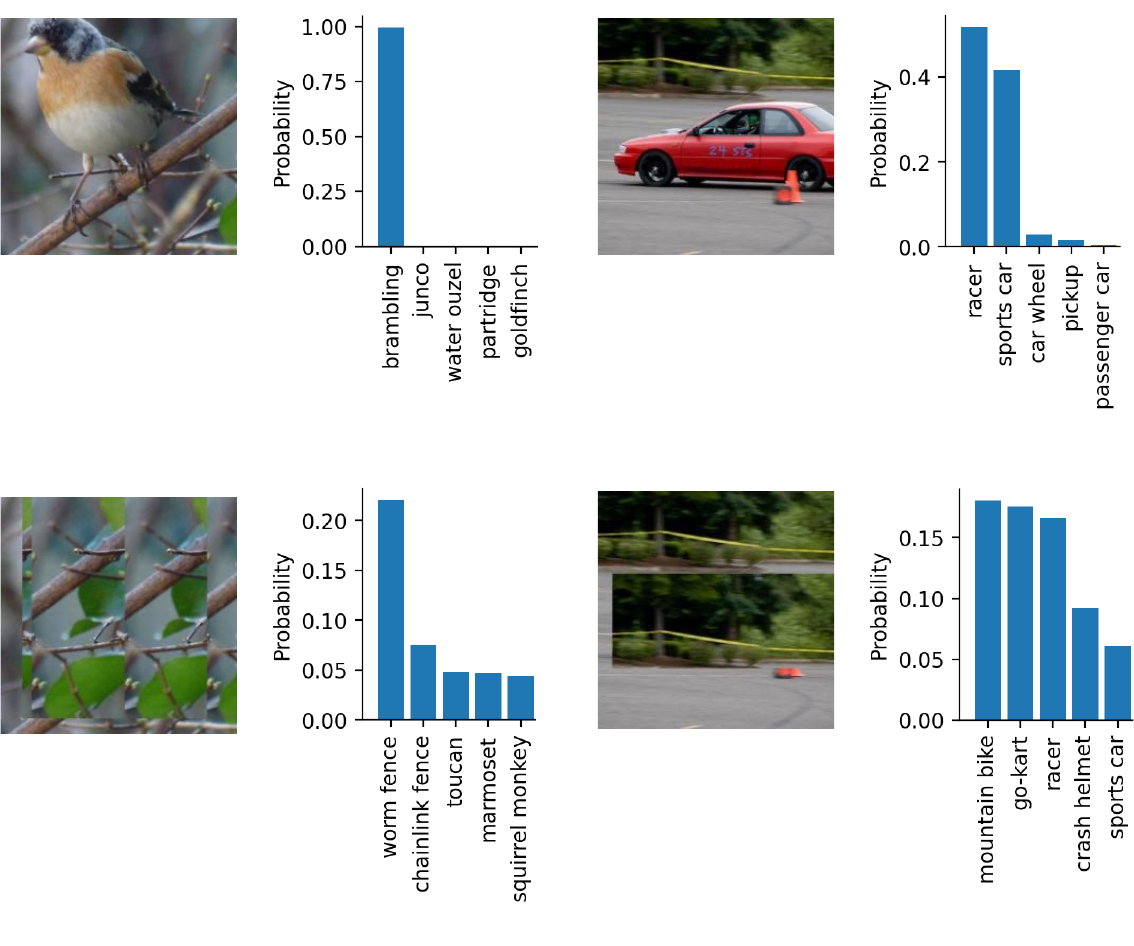}
     \caption{Top-5 class prediction probabilities on the original (top) and their corresponding Only-BG-T (bottom) samples from the ImageNet-9 dataset.}
     
     \label{fig:bg_ch_sp}
 \end{figure}

To further investigate this observation, we conducted a detailed analysis using the Only-BG-T configuration from the dataset. In this configuration, the foreground is obscured with a portion of the background taken from the same image. As shown in Figure~\ref{fig:bg_ch_sp} (top), when the original images of a car and a bird are processed through the image classifier, the predicted ImageNet classes are mostly relevant to their respective categories. However, when the foreground is concealed, as illustrated in Figure~\ref{fig:bg_ch_sp} (bottom), the classifier still predicts \texttt{bird} and \texttt{wheeled vehicle} types. These findings corroborate the observations made in Figure~\ref{fig:bg_ch_wc}. For instance, \methodname{} successfully detects "tree" and "branch" as dominant features for the \texttt{bird} category. When the bird is concealed, as shown in Figure~\ref{fig:bg_ch_sp} (bottom), the classifier tends to associate the remaining "branches" with the concept of a bird. Similarly, the car example in Figure~\ref{fig:bg_ch_sp} shows that "street" and "road" are correctly identified by \methodname{} in Figure~\ref{fig:bg_ch_wc}.

\methodname{} not only exposes classifier bias, but it also has the potential to reveal dataset properties. For instance, in Figure~\ref{fig:bg_ch_sp}, the prevalence of the \texttt{man} category for \texttt{fish} suggests that the dataset may contain many images of fishermen displaying their catches. Similarly, comparing the \texttt{dog} categories in the Background Challenge dataset indicates that the former likely has more outdoor images of dogs playing, as evidenced by the presence of grass and Frisbee, while the latter has more indoor images of dogs, indicated by the prominence of beds. Therefore, \methodname{} can serve as a tool for detecting bias in datasets and may provide insights on how to mitigate such biases by including samples from underrepresented classes to achieve balance. 

\begin{figure}[!t]
     \centering
     \includegraphics[width=0.8\textwidth]{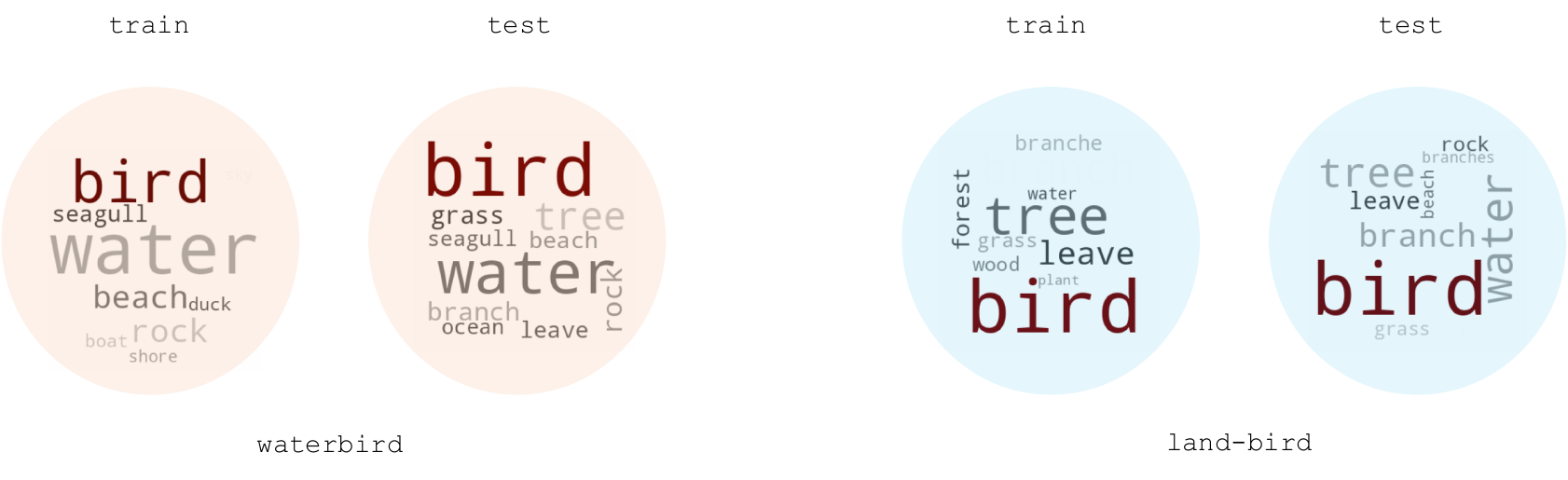}
     \caption{Word clouds of the explanations generated by \methodname{} for the Waterbirds dataset. \methodname{} reveals significant feature shifts between the training and test sets within each category. Notably, in the training set of \texttt{waterbirds}, the attribute "water" outweighs "bird" in dominance compared to its respective test set. Additionally, the test set of \texttt{landbirds} exhibits the presence of "water" and "beach" attributes, which are absent in its corresponding training set.}
     
     \label{fig:wtr_bird}
 \end{figure}

\textbf{Waterbirds.} The Waterbirds dataset, introduced by Sagawa et al.~\citep{sagawa2019distributionally}, serves as a benchmark for evaluating the extent to which models capture spurious correlations present in the training set. We employed \methodname{} to analyze both the training and test sets of each category, specifically the \texttt{waterbirds} and \texttt{landbirds} categories. As depicted in Figure~\ref{fig:wtr_bird}, \methodname{} successfully identified significant shifts in the feature spaces between the training and test sets of each category. Notably, in the training set of the \texttt{waterbirds} class, the attribute "water" exhibited a much stronger presence compared to "bird", whereas in the \texttt{waterbirds} test set, these two features were more balanced. Additionally, in the test set of \texttt{waterbirds}, \methodname{} detected land attributes such as "grass", "tree", and "branch", which were not prominent in the training set of \texttt{waterbirds}. In essence, the \texttt{waterbirds} test set contained images of waterbirds on land, which the model had not encountered during training. It is worth noting that this subgroup (waterbirds on land) represents a particularly challenging category for most trained classifiers, as their performance tends to be subpar in such cases. Similarly, when examining the word clouds of the \texttt{landbirds} class, the attribute "water" became dominant in the test set compared to the training set of \texttt{landbirds}, where \methodname{} did not identify such an attribute.

\textbf{Leveraging \methodname{} to Mitigate Spurious Correlations.} In addition to visualizing and detecting potential spurious correlations and shortcuts learned by a model, \methodname{} can be utilized to enhance the model's performance when encountering such correlations. In the case of the Waterbirds dataset, although most classifiers exhibit reasonable overall accuracy on the test set, their performance significantly deteriorates when evaluating specific subgroups. Our objective is to improve accuracy specifically for the worst-performing subgroup while maintaining a high average accuracy. To achieve this, we train classifier on the Waterbirds dataset and  employ \methodname{} on the training set to identify the "problematic" samples.
It is our assumption that the samples in which a class with a prominent presence other than the correct one is observed, are the instances in which the classifier is giving undue attention to the spurious features.
Therefore, we select samples whose the first dominant feature is something other than "bird". By pinpointing these samples ($29\%$ of the training data), as done in~\citep{asgari2022masktune}, we utilize GradCAM~\citep{selvaraju2017grad} to localize and mask irrelevant areas (the first non-bird dominant feature that \methodname{} identifies) in the input. Subsequently, we fine-tune the trained model exclusively using the masked samples. The results presented in Table~\ref{tab:waterbirds} demonstrate that employing \methodname{} yields a substantial improvement in the accuracy of the worst-performing subgroup while maintaining the average accuracy. Furthermore, in the second and third rows of the table, we compare this approach to randomly masking the training set, to match the number of samples identified as problematic by \methodname{}.

\renewcommand{\arraystretch}{1.1}
\begin{table*}[h!]
\centering
\caption{Classification results from the Waterbirds dataset using ViT. Our method significantly improves empirical risk minimization (ERM)'s accuracy on the worst-group. Results are averaged over three different runs. In each run, \methodname{} demonstrated a substantial improvement in the ERM performance while outperforming other methods using significantly fewer masked samples during fine-tuning.}
\begin{tabular}{lcccc}
\cline{2-5}
                & ERM             & RandMask                 & MaskTune        & \methodname{} (ours)         \\ \hline
Subgroup-1              & \textbf{98.20 $\pm$ 0.72}   & 98.14 $\pm$ 1.76         & 97.63 $\pm$ 2.22 & 97.96 $\pm$ 0.25           \\
Subgroup-2               & 89.49 $\pm$ 5.35 & 90.10 $\pm$ 2.80           & 93.62 $\pm$ 5.11 & \textbf{94.06 $\pm$ 1.82}  \\
Subgroup-3               & 97.95 $\pm$ 1.15 & 98.03 $\pm$ 1.97           & 97.96 $\pm$ 2.25 & \textbf{98.12 $\pm$ 0.47}  \\
Worst-group              & 89.79 $\pm$ 8.13 & 91.72 $\pm$ 2.49          & 95.195 $\pm$ 5.33 & \textbf{95.63 $\pm$ 2.55} \\
Average             & 96.34 $\pm$ 2.12 & 96.59 $\pm$ 0.91           & 97.10 $\pm$ 0.71 & \textbf{97.39 $\pm$0.16}            \\ 
Masked Samples             & N/A & 100\%           & 100\%
& \textbf{26\%}            \\ \hline

\end{tabular}

\label{tab:waterbirds}
\end{table*}

\section{Related Work}
\paragraph{Visual Heat Map-based Explanations.}
A significant body of research has focused on post-hoc explanation techniques for image classifiers, including methods such as Grad-CAM~\citep{selvaraju2017grad}, LIME~\citep{ribeiro2016should}, CAM~\citep{wang2020score}, ablation studies~\citep{ramaswamy2020ablation}, DeepLIFT~\citep{collins2018deep}, and saliency maps~\citep{fong2017interpretable}. These methods typically rely on network gradients or perturbation analysis to generate heat maps that highlight the most relevant regions in an input image for the classifier's decision. While these approaches effectively indicate the areas contributing to the classifier's prediction, they lack the ability to provide a detailed understanding of the specific features learned by the model. Moreover, interpreting these heat maps can often be challenging and subjective. In contrast, our proposed approach leverages textual explanations to represent the learned features captured by the classifier, offering a more intuitive and direct interpretation of its decision-making process. By visualizing the dominant words, our method provides a comprehensive and accessible means to comprehend the underlying features encoded by the classifier, enabling a deeper understanding of its behavior and facilitating more informed analysis.

\paragraph{Textual Explanation of Vision Models.}
Previous research has demonstrated the efficacy of incorporating textual explanations in training vision models, particularly in the context of multi-modal setups like visual question answering~\citep{park2018multimodal,sammani2022nlx}. Furthermore, the utilization of large-scale vision-language models in classification tasks has shown promising self-explanatory capabilities~\citep{radford2021learning,li2022blip,li2023blip,jia2021scaling,singh2022flava}. Notably,~\cite{menon2022visual} recently proposed a technique to improve the interpretability of vision-language models used for image classification. However, the existing studies predominantly concentrate on elucidating vision-language models trained and fine-tuned jointly. There might be \textit{architectural} similarities between our work and recent concurrent vision-language models such as~\citep{li2023blip, zhu2023minigpt, liu2023visual}. However, it is important to note that these models are designed for a different objective, such as image captioning, where a vision and a language model are trained together or individually. The exploration of interpreting the frozen embeddings of independently trained image classifiers using trained (and frozen) language models remains largely unexplored or deficient in current methodologies. This gap underscores the need for novel approaches that specifically address the challenge of interpreting independently trained image classifiers—an aspect that our proposed method aims to tackle.

\section{Conclusion}
We presented \methodname{}, a novel method that leverages the power of language models to interpret the learned features of independently trained image classifiers. Our approach enabled the generation of comprehensive textual explanations for learned visual features, revealing spurious correlations, biases, and uncovering underlying patterns. To validate the efficacy of \methodname{}, we conducted a series of experiments to ensure its proper functioning and reliability. We then demonstrated a practical application of the explanations that \methodname{} generates in identifying and mitigating spurious correlations ingrained within image classifiers. We uncovered and addressed these undesirable correlations, thereby enhancing the reliability and accuracy of the classifiers. This highlights the potential of \methodname{} as a valuable tool for finding and combating spurious correlations to promote more robust and trustworthy image classification models.

\methodname{} offers multiple applications in the realm of understanding and analyzing classifiers. Firstly, it can be employed to gain insights into the specific features that have been learned by a classifier. By generating comprehensive explanations, \methodname{} allows us to delve into the inner workings of the model and comprehend the learned representations. Secondly, it serves as a valuable tool for identifying biases and spurious correlations within the classifier. This capability enables the detection and mitigation of undesirable shortcuts or unintended associations that the model may have picked up during training. Lastly, \methodname{} can be utilized to assess data bias, assuming that the trained model itself is unbiased. By examining the generated explanations, we can gain valuable insights into any underlying biases present within the dataset being processed by the model. Overall, \methodname{} offers a versatile framework for uncovering and addressing various aspects of interpretability, bias, and feature analysis within classifiers.

In this paper we showed the application of \methodname{} on image classifiers and stable diffusion. An interesting avenue for future research is to explore its potential in other domains such as image segmentation and auto-encoders and other other architectures and input modalities. By adapting and applying \methodname{} to these contexts, we can gain valuable insights into the learned representations and underlying concepts within these models. Furthermore, it would be valuable to expand \methodname{} to encompass other data types, particularly in the realm of 3D (pointcloud, voxel, etc.) classifiers. One commonly held belief is that 3D data can inherently lead to capturing geometric information. Therefore, leveraging \methodname{} to investigate whether 3D encoders indeed extract geometry-related features could provide valuable insights into the learning process of these models. In this work, we performed multiple checks to ensure the explanation is indeed relevant to the learned visual features. However, we encourage using the method with even more checks to ensure that potentially learned biases from language models do not contribute to the final explanation.

\bibliography{refs}

\end{document}